\documentclass[journal]{IEEEtran}
\usepackage{graphicx} 
\usepackage{caption}
\usepackage{cite}
\usepackage[numbers,sort&compress]{natbib}
\usepackage{amssymb}
\usepackage{subcaption}
\usepackage{multirow}
\usepackage{listings}
\usepackage{xcolor}
\usepackage{booktabs}
\usepackage[framemethod=TikZ]{mdframed}
\usepackage{svg}

\begin{document}
\title{RobotGPT: Robot Manipulation Learning from ChatGPT}

\author{Yixiang Jin $^{1}$, Dingzhe Li $^{1}$, Yong A $^{1}$, Jun Shi $^{1}$, Peng Hao $^{1}$, Fuchun Sun $^{2}$, Jianwei Zhang $^{3}$, Bin Fang $^{2*}$

\thanks{This work was supported by Major Project of the New Generation of Artificial Intelligence, China (No. 2018AAA0102900), the National Natural Science Foundation of China under Grant 62173197}
\thanks{$^{1}$The authors are with Samsung Research China – Beijing (SRC-B)
        {\tt\small \{yixiang.j, dingzhe.li, yong.a, jun7.shi, peng1.hao
\} @samsung.com}}
        
\thanks{$^{2}$The author is with Tsinghua University
        {\tt\small \{fangbin, fcsun\}@tsinghua.edu.cn}, *Bin Fang is the corresponding authors.}%

\thanks{$^{3}$The author is with Universität Hamburg, Germany}%

}

\markboth{IEEE Robotics and Automation Letters}{\hfill My Right Header}

\maketitle
\pagestyle{empty}  
\thispagestyle{empty} 

\begin{abstract}
We present RobotGPT, an innovative decision framework for robotic manipulation that prioritizes stability and safety. The execution code generated by ChatGPT cannot guarantee the stability and safety of the system. ChatGPT may provide different answers for the same task, leading to unpredictability. This instability prevents the direct integration of ChatGPT into the robot manipulation loop. Although setting the temperature to 0 can generate more consistent outputs, it may cause ChatGPT to lose diversity and creativity. Our objective is to leverage ChatGPT's problem-solving capabilities in robot manipulation and train a reliable agent. The framework includes an effective prompt structure and a robust learning model. Additionally, we introduce a metric for measuring task difficulty to evaluate ChatGPT's performance in robot manipulation. Furthermore, we evaluate RobotGPT in both simulation and real-world environments. Compared to directly using ChatGPT to generate code, our framework significantly improves task success rates, with an average increase from 38.5\% to 91.5\%. Therefore, training a RobotGPT by utilizing ChatGPT as an expert is a more stable approach compared to directly using ChatGPT as a task planner.
\end{abstract}

\section{Introduction}
Large language models (LLMs) have demonstrated impressive achievements across various tasks, including but not limited to text generation, machine translation, and code synthesis. Recently, there has been a growing trend of work  \cite{vemprala2023chatgpt, lin2023text2motion} attempting to incorporate LLMs into robotics systems. This previous work has demonstrated that LLMs are capable of conducting robot system planning in a zero-shot fashion. However, to date, no research explores the full extent of what tasks LLMs are capable of solving.

In addition, the rapid advancement in LLMs has also a significant impact on Human-Robot Interaction (HRI). Research on HRI involves broader areas, such as virtual reality, smart spaces, and more. To increase user-facing popularity and to make it more comfortable and convenient for people to use robots, natural interaction is one of the key technologies. At present, some progress has been made in the research of natural language interaction \cite{marge2022spoken}, but there are also many shortcomings.

Nevertheless, there is still a slight lack of generalization of usage scenarios and comprehension of language models. The recent advent of ChatGPT has raised expectations for LLMs. ChatGPT stands out among various language models due to its powerful code model generation capabilities and conversational flexibility, showing an amazing understanding that allows users to interact with robots in a more natural way. Existing representative works use ChatGPT to generate code that can be deployed directly on robots, describing the goals of the task and the library of functions available to ChatGPT in advance, and then using ChatGPT to manipulate different robots to perform functions such as drone navigation and robots grasping objects in natural language after continuous cyclic feedback. The ability of ChatGPT to parse user intent from natural conversations and to generate problem-solving code from natural conversations reveals the great potential of ChatGPT for applications in robotics \cite{vemprala2023chatgpt}. 

In this paper, we conducted in-depth research on the application of ChatGPT in the field of robot manipulation. We aim to advance the practical application of ChatGPT in robotics. With ChatGPT, we implement a framework that translates the environment and tasks into natural language. Subsequently, ChatGPT generates specific action command codes, which are used to train the agent RobotGPT that leverages ChatGPT's problem-solving capabilities. The robot can indirectly use natural language to interact with the outside world in order to design reasonable action sequences and implement corresponding functions (e.g., pick-and-place). We also have some suggestions for dialogue with ChatGPT, i.e., how to structure prompts so that ChatGPT can understand more accurately and deeply. In addition, we point out the limitations and security risks of such interactions, as well as simple ideas to solve the problem. Overall, our main contributions are as follows:

\begin{enumerate}

\item We explore an effective prompts structure with a self-correction module, and conducted experiments on tasks of varying difficulty to explore the capability boundaries of ChatGPT on robotic tasks.

\item We propose a novel framework for the application of ChatGPT in the field of robotics. Our system does not directly execute the code generated by ChatGPT. Instead, we employ an agent that learns the planning strategies generated by ChatGPT, thereby enhancing the stability of the system. Although fine-turning can improve accuracy \cite{liang2022code}, training data is difficult to obtain.

\end{enumerate}

\begin{figure*}[h!t]
\centering
\includegraphics[width=\linewidth]{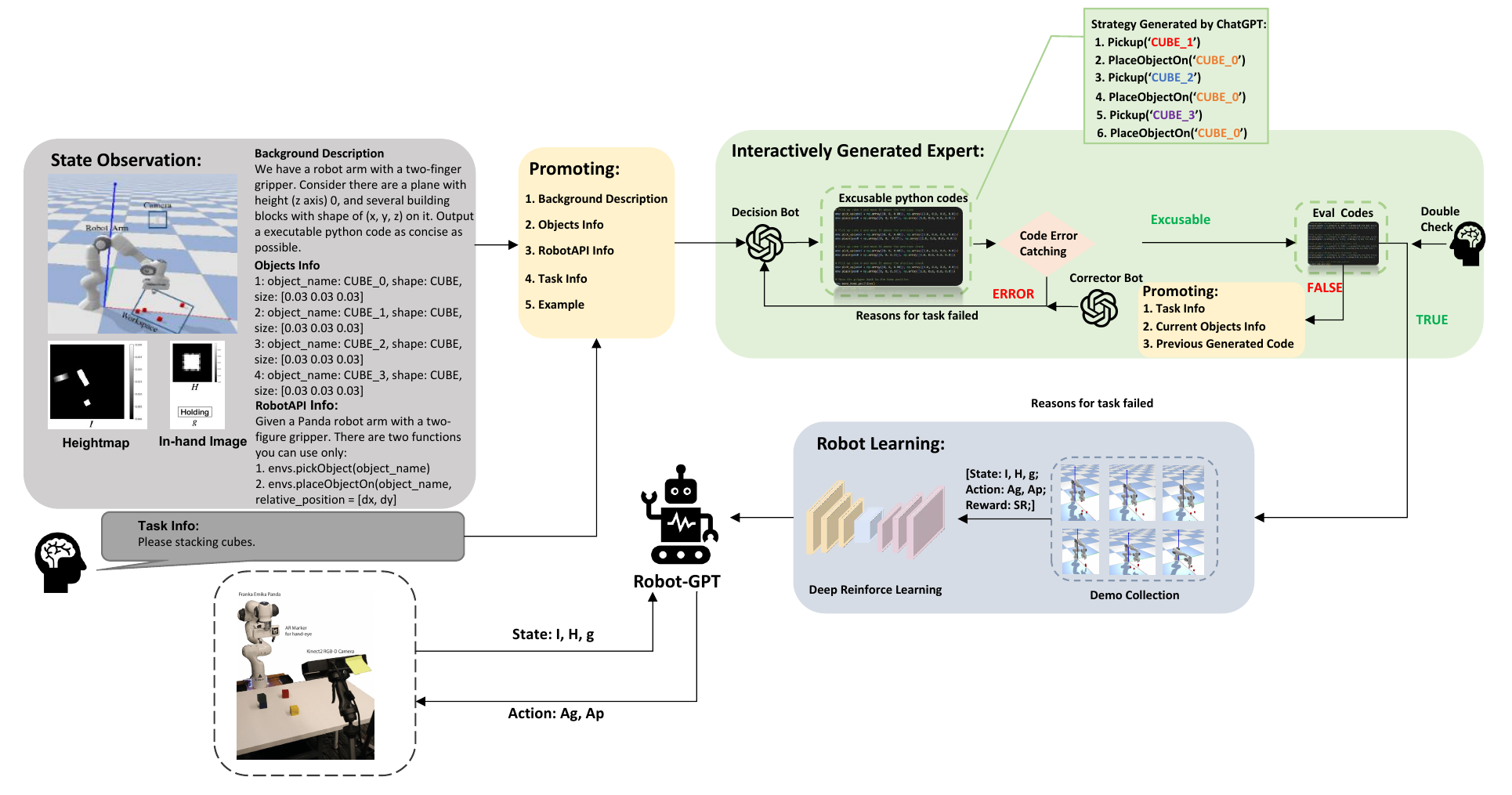}
\caption{Architecture of our system. ChatGPT plays three roles within it, namely decision bot, evaluation bot and corrector bot. The operator gives an instruction for the robot to complete the task, and then a natural language prompt is generated based on environmental information and human instruction. 
The decision bot will generate the corresponding executable code based on the provided prompts. Next, the generated code will be executed line by line. If a runtime error occurs, the reason for this error and the line of code where the error occurred will be provided for the decision bot for modification until the code can run successfully. Then, the executable code will be tested by the Eval Code model that is generated by the evaluation bot. If the executable code can’t pass the Eval Code, the corrector bot will analyze potential reasons for the failure of results and send those failure reasons back to the decision bot for correction.
Afterward, the code that satisfied the evaluation condition will be used to generate demonstration data. After training, the trained agent can deploy the real robot perfectly.}
\label{figure_framework}
\end{figure*}

\section{Related Work}
\subsubsection{LLMs for robotics.} Controlling robots via language can bring more natural interaction for non-experts \cite{tellex2020robots}. Stability and generalization are concerns for using language to control robots \cite{liang2022code}. There exists a large literature on this issue, overall divided into high-level interpretation (e.g., semantic parsing, planning) \cite{kollar2010toward} and low-level policies (e.g., model-based, imitation learning or reinforcement learning)  \cite{nair2022learning}. Large language models (LLMs) exhibit powerful general intelligence capabilities. At the same time, LLMs for robots have received considerable attention \cite{madaan2022language}. The typical literature related to our research work is as follows. \cite{huang2022language} proposes LLMs can effectively decompose high-level tasks into mid-level plans without any further training. \cite{lin2023text2motion} construct geometrically feasible plans based on LLMs. \cite{brohan2023can} use value functions to evaluate each step of LLMs generation and select the optimal trajectory. \cite{wu2023tidybot} build systems based on LLMs to learn the particular person's preferences. \cite{liang2022code} use LLMs to generate robot-centric programs. However, the stability of the output from LLMs is still worth exploring.
\subsubsection{Robot Learning.} 
In order to exploit the ability of robots to interact with the real world, robot learning has become a research hotspot \cite{kroemer2021review}. Deep Learning advances the development of robot learning when the state includes image \cite{cabi2019scaling}. There are many algorithms for robot learning. However, the algorithms based on reinforcement learning and imitation learning are still the mainstream \cite{wang2022equivariant,zeng2022robotic}. In order to compare different reinforcement learning algorithms, many benchmarks have been proposed. \cite{ lee2021ikea} are targeted at single tasks (e.g., door opening, furniture assembly, in-hand dexterous manipulation). \cite{zhu2020robosuite, delhaisse2020pyrobolearn} have a variety of different environments, but lack long-horizon tasks. The relatively comprehensive benchmarks are RLBench \cite{james2020rlbench} and BulletArm \cite{wang2023bulletarm}. Although benchmarks provide the framework, how to get data for robot learning is still a problem.

\section{Methodology}

ChatGPT cannot parse visual inputs and operate robots by themselves. In this paper, we leverage a simulation environment and natural language-based robotic API to unleash ChatGPT's general problem-solving capabilities. As a result, we expect that the trained agent RobotGPT can absorb ChatGPT's knowledge at the task planning level.

\begin{figure*}[htp]
\centering
\includegraphics[width=\linewidth]{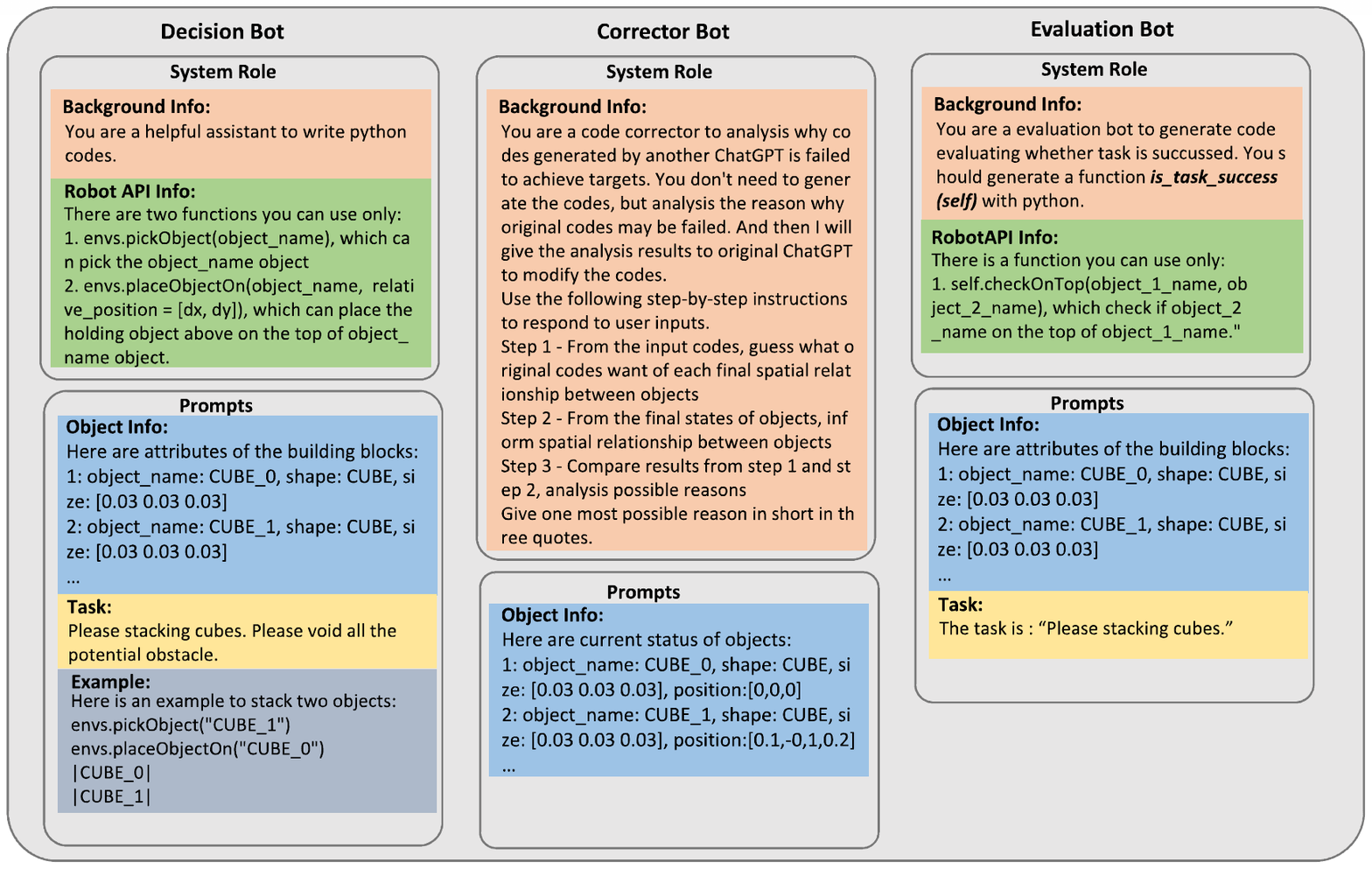}
\caption{Prompts for ChatGPT}
\label{prompts}
\end{figure*}

\subsection{ChatGPT prompts for robot manipulation}

Recently, there has been a growing interest in using large language models such as ChatGPT to directly control robots by generating high-level actions. However, this approach may not be suitable for several reasons. Firstly, ChatGPT-generated actions may not be safe or stable, as they do not take into account the physical constraints and limitations of the robot. Secondly, ChatGPT lacks the ability to reason about causal relationships and temporal dependencies, which are crucial for controlling complex robotic systems. Therefore, we propose an alternative approach based on robot learning, where ChatGPT is used to generate demonstrations that are used to train the robot. By leveraging the strengths of both language models and robot learning, we aim to develop safer and more robust robotic systems. In this section, we detail our framework of interaction with ChatGPT for demonstration generation. 

We propose a framework for interacting with ChatGPT that consists of two parts: code generation and error correction. In the code generation phase, the user describes the task and provides examples to guide ChatGPT's response. This helps to ensure that ChatGPT generates appropriate and relevant outputs that meet the user's requirements. In the error correction phase, both runtime errors and task failures are considered to be correct.

\subsubsection{Prompting description}
Effective prompting methods are essential for improving the performance of ChatGPT in various domains. Vemprala et al. \cite{vemprala2023chatgpt} pointed out the current challenges of prompting LLMs for robotic manipulation as: 1). requiring a complete and accurate description of the problem; 2). the allowable natural language described APIs; 3). biasing the answer structure. In this section, we detail our effective prompts method for robotic manipulation. We propose a five-part prompting method that includes background description, object info, environment info, task info, and examples. In the background description part, the basic information about the environment is described, such as the purpose of the environment, its layout, and relevant entities. In the object info part, we list all objects' names, shapes, poses, and other helpful information, such as their properties and relationships with other objects. In the environment info part, we describe the robot and API functions ChatGPT can use to perform the task. In the task info part, we give the specific task for ChatGPT, generally to generate Python code for a given job. Finally, in the example part, we provide some examples to facilitate a better understanding of the environment and API usage. Following the suggestion by OpenAI \cite{openai}, we set background information and RobotAPI information as the system message in the ChatGPT API to obtain satisfactory responses. By using this comprehensive and structured prompting method, we aim to improve the accuracy and efficiency of ChatGPT in various tasks and domains. 


\subsubsection{Self-correction}
In generating responses for complex tasks, ChatGPT may occasionally produce minor bugs or syntax errors that necessitate correction. This paper introduces an interactive approach for rectifying ChatGPT's responses. To employ this method, we first execute the generated code within a simulator and assess the outcomes. 

The generated code will be executed line by line, and when a runtime error occurs, the runtime errors, including the error message and its location, will be captured by the Code Error Catching module. This data is then sent back to the ChatGPT decision bot for further analysis. In situations where the result is a failure, the corrector bot can analyze potential reasons for the failure based on prompts and generate a response explaining why the task failed. Finally, the original ChatGPT decision bot will regenerate the code based on the corrector bot’s failure analysis. Utilizing this feedback, ChatGPT amends its response and produces accurate code. This interactive process may iterate up to three times. Our objective is to improve the precision and dependability of ChatGPT's responses, making them increasingly relevant across a range of domains. 

\subsubsection{Generated Code Evaluation}
The task completed according to the ChatGPT-generated code should satisfy the requirements. To this end, an automatic, efficient and precise task evaluation module is imperative. 

As Fig. \ref{figure_framework} shown we employ a ChatGPT named evaluation bot to generate evaluation code. The prompts for the evaluation bot have some differences with the decision bot. The structure of prompts remains unchanged, but the content of background description and robot API has been different as Fig. \ref{prompts} presented. Evaluation bot generated function is\_task\_success() will serve as the criterion for determining the success of the entire task. The role of humans is to double-check whether the generated evaluation code is correct. If an incorrect evaluation code is generated, humans will intervene to make corrections. This kind of design can minimize the burden on humans.




\subsection{Robot learning}
It is unreliable to rely on ChatGPT to perform general robotics tasks because the output of ChatGPT is random, which exaggerates the risk of robotic work. Although setting the temperature to zero can produce consistent outcomes at the cost of diminishing diversity and creativity, it may also lead to the continual failure of tasks. To solve this problem, we expect robots to learn robot policies to absorb ChatGPT's knowledge of solving general tasks. For robot learning, we leverage the state-of-the-art, open-source robotic manipulation benchmark and learning framework BulletArm \cite{wang2023bulletarm} to train an agent from a ChatGPT-generated demonstration.  
\subsubsection{Action, state space and reward}
The ChatGPT-powered expert demonstrations were generated in the simulation environment, which comprised a Panda robot with a camera mounted on top of the workspace. The state space is composed of a top-down height-map $H$, an in-hand image $I$, and gripper state $g \in \{HOLDING, EMPTY\}$. For action space, it included robot skill $A_s \in \{PICK, PLACE\}$ and target pose $A_p \in A^{xy\theta}$. Where x and y represent the XY coordinates of the end-effector and $\theta$ notes rotation along the z-axis. The reward is set as a sparse reward function, in which the reward is 1 when all state-action pairs reach the goal state and 0 otherwise.  
\subsubsection{Algorithm}
\begin{figure}
\centering
\includegraphics{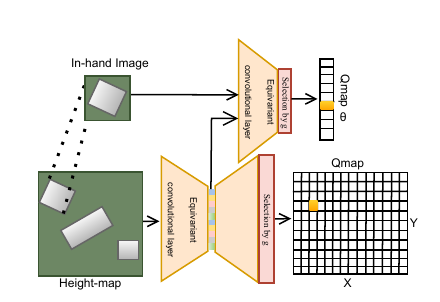}
\caption{Robot learning network architecture} 
\label{fig_network}
\end{figure}

 BulletArm\cite{wang2023bulletarm} shows that SDQfD \cite{wang2020policy} performs better than DQN \cite{mnih2015human}, ADET \cite{lakshminarayanan2016reinforcement} and DQfD \cite{hester2018deep}. As for network architecture, Equivariant ASR\cite{wang2022equivariant} has the best performance across all environments, then Rot FCN\cite{zeng2018learning} and CNN ASR\cite{wang2020policy}, and finally FCN\cite{long2015fully}. In this paper, the SDQfD\cite{wang2020policy} algorithm is adopted for the task of robot learning with Equivariant ASR network\cite{wang2022equivariant} as Fig. \ref{fig_network} shown. The loss function is the sum of n-step TD loss and strict large margin loss.

\section{Experiments}
In this section, we conduct the evaluation of the proposed system in both simulation and real environments. To be more precise, we focus on explaining the following question:

\begin{enumerate}

\item Can Robot-GPT efficiently and safely collects demonstration data and deploy it in the real world bridging the sim-to-real domain gap?

\item Can our LLM-driven robot solves problems that hand-coding and non-LLM are not well-addressed?

\end{enumerate}

\begin{figure}[h]
\centering
\includegraphics[width=\linewidth]{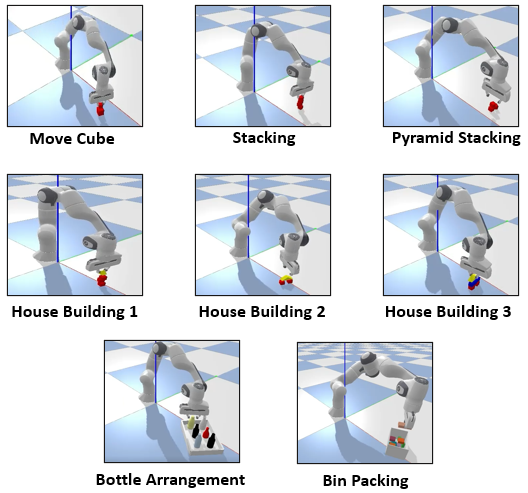}
\caption{Eight tasks used in our experiments}
\label{fig_tasks}
\end{figure}

\begin{table}[h]
\centering
\caption{Inference Comparison}
\begin{tabular}{p{2cm}p{5cm}}
\toprule
\textbf{Task Name} & \textbf{Description}\\
\midrule
move\_cube & Move small cube above onto big cube \\
\midrule
stacking & Stack the given blocks together. \\
\midrule
pyramid\_stacking & Stack the given three blocks into a pyramid shape.\\
\midrule
house\_building\_1 & Construct a tall building using the given three blocks and a triangle shape.\\
\midrule
house\_building\_2 & Construct a bungalow using the given two cubes and a triangle shape.\\
\midrule
house\_building\_3 & Construct a house using the given two cubes  (red), a brick (blue) and a triangle shape.\\
\midrule
bottle\_arrangement & Arrange the given six bottles neatly on a tray.\\
\midrule
bin\_packing & Pick up blocks on the table and place on tray.\\
\bottomrule
\label{tab:inf}
\end{tabular}
\end{table}

\subsection{Metrics}
To create a grading system, we consider the following three aspects: the number of objects $o$, object categories $c$ and the number of tasks' steps $s$. These three factors are the top influences on the difficulty of robotic desktop grasping tasks, as determined through a survey conducted with 32 experts and engineers in the field of robotics and computer vision. The questionnaire consists of two main parts. The first part is to write down the top three factors influencing robot manipulation difficulty. The second part is to assign a score to eight scenarios for this experiment. Results are shown in TABLE \ref{tab:diff}.

Among the three factors, the number of objects has the greatest impact, and we have magnified its weight in the score. Therefore, the score of task difficulty can be calculated using the following equation:
\begin{equation}
\label{eq1}
score = o + o * c + s
\end{equation}

Tasks with scores between 0 and 10 are considered easy, tasks with scores between 11 and 20 are regarded as medium, and scores above 20 are defined as difficult tasks. Table \ref{tab:diff} shows the tasks utilized in the experiment and their corresponding difficulty levels. The difficulty calculated from equation \ref{eq1} is the same as the subjective results obtained from the survey questionnaire besides bin\_packing task, which indicates that this evaluation system has general applicability. The reason for the subjective result of bin\_packing task leading to simple results is due to ignoring optimizing the placement location to fill the bin without objects falling.

\begin{table}[h!t]
\centering
\caption{Experiment tasks and their corresponding difficulty. ‘Difficulty1’ represents results calculated based on the metrics, and ‘Difficulty2’ is subjective ratings obtained from the survey responses.}
\label{tab:diff}
\begin{tabular}{|c|c|c|c|c|c|c|}
\hline
\textbf{Task Name} & \textbf{$o$} & \textbf{$c$} & \textbf{$s$} & \textbf{Score} & \textbf{Difficulty1}& \textbf{Difficulty2}\\ \hline
move cube& 2 &  1 & 2 & 6 & E & E\\ \hline
block stacking & 4 & 1 & 6 & 16 & M & M \\ \hline
pyramid stacking & 3 & 1 & 6 & 15 & M & M\\ \hline
house building 1 & 4 & 2 & 6 & 20 & M & M\\ \hline
house building 2 & 3 & 2 & 4 & 15 & M & M\\ \hline
house building 3 & 4 & 3 & 6 & 28 & H & H\\ \hline
bottle arrangement & 6 & 1 & 12 & 30 & H & H \\ \hline
bin packing & 8 & 1 & 16 & 40 & H & E\\ \hline
\end{tabular}
\end{table}

In the following quantitative experiment, we will generate 25 random scenes for each task and count the number of successful attempts.

\subsection{Experiment setup}
\begin{figure}[h!]
  \centering
  \begin{subfigure}[b]{0.49\linewidth}
      \includegraphics[width=\linewidth]{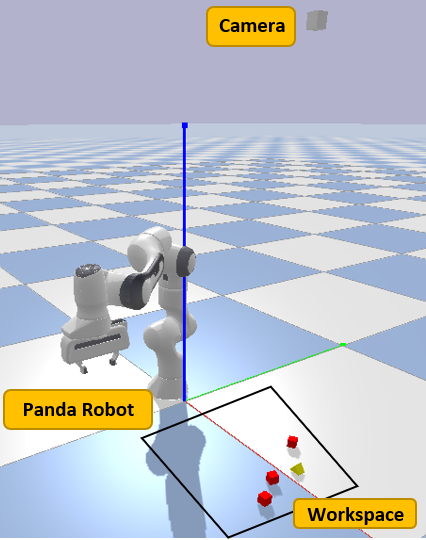}
      \caption{Simulation Environment}
      \end{subfigure}
  \hfill
  \begin{subfigure}[b]{0.49\linewidth}
      \includegraphics[width=\linewidth]{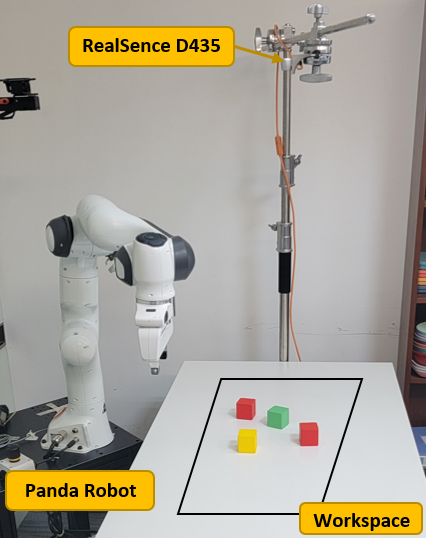}
      \caption{Real robot environment}
  \end{subfigure}
  \caption{Experiment Setup}
  \label{fig:setup}
\end{figure}

Fig. \ref{fig:setup} shows our experiment setup in both simulation and real environments. We mount an RGB-D sensor directly above the workspace to provide a clear height map of the scene. In the simulated environment, the robot relies on the PyBullet engine for motion control. While in the real world, the robot utilizes MoveIt and ros\_franka for motion planning and execution.

\subsection{Simulation experiment}
Table \ref{table_result} presents the quantitative results of the eight experiments. The fact is that despite entering the same prompts each time, the generated code and the resulting output always have significant differences because the temperature of the decision bot is 1.0. In addition, the code generated by ChatGPT contains syntax or logic errors. Although our self-correction module can revise some syntax errors, in most cases, if ChatGPT fails to generate successful code initially, it becomes difficult to achieve success in this experiment. 
\begin{figure}[h!]
  \centering
  \begin{subfigure}[b]{0.3\linewidth}
       \includegraphics[width=\linewidth]{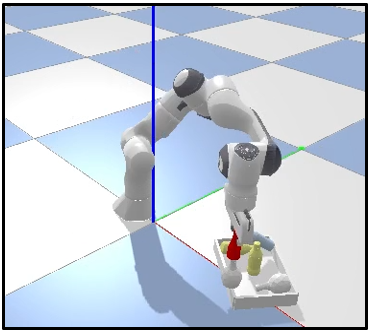}
      \caption{Failing to understand requirements}
      \end{subfigure}
  \hfill
    \begin{subfigure}[b]{0.3\linewidth}
      \includegraphics[width=\linewidth]{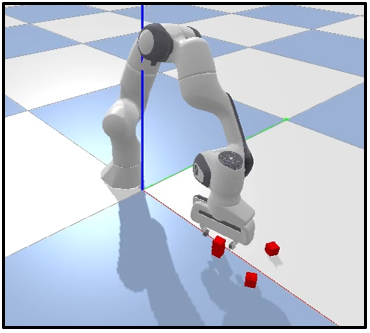}
      \caption{Wrong action sequence planning}
      \end{subfigure}
  \hfill
  \begin{subfigure}[b]{0.3\linewidth}
      \includegraphics[width=\linewidth]{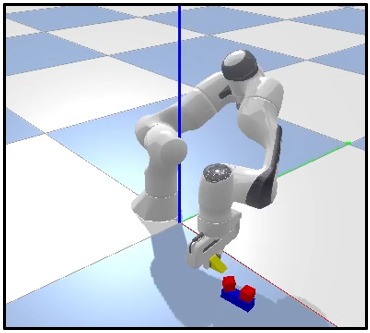}
      \caption{Wrong placement position}
  \end{subfigure}
  \caption{Failed task planning generated by ChatGPT.}
  \label{fig_failure}
\end{figure}

 Fig. \ref{fig_failure} illustrates the three most common failures caused by incorrect planning generated by ChatGPT. Fig. \ref{fig_failure} (a) shows failures caused by misunderstanding of the task requirements. The bottle\_arrangement task requires placing the bottles neatly on the tray, rather than dropping them onto the tray at will. Fig. \ref{fig_failure} (b) presents wrong action sequence planning, in which the robot is grasping the stacked blocks in the image. This is obviously unreasonable because the robot should be grasping objects that have not yet been stacked. In Fig. \ref{fig_failure} (c), it is evident that the placement position of the robot is deviant. Therefore, ChatGPT can provide different solutions for the same prompts, some of which are correct while others are incorrect. This is why we propose the RobotGPT framework as a more stable approach.
\begin{figure}[h]
\centering
\includegraphics[width=\linewidth]{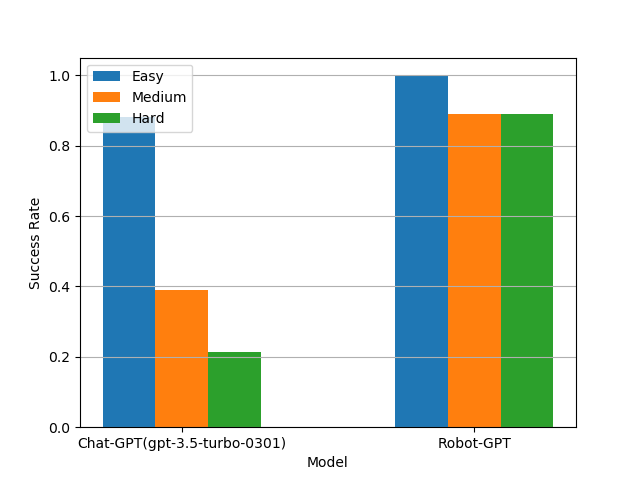}
\caption{Success rates for three difficulty levels}
\label{fig_result}
\end{figure}

Fig. \ref{fig_result} displays the success rates for three difficulty levels. For ChatGPT, it is evident that as the task difficulty increases, the success rate decreases significantly. The success rates for the easy, medium, and difficult tasks are 0.88, 0.39, and 0.21, respectively. In contrast, our RobotGPT model demonstrates robustness across all levels of tasks, maintaining a good performance, which can achieve 0.915 on average. 

\begin{table}[h!t]
\centering
\caption{The counting results of experiments using a gpt-3.5-turbo-0301 model}
\begin{tabular}{|c|c|c|c|c|c|c|}
\hline
\multirow{2}{*}{\textbf{Task Name}} & \multicolumn{3}{|c|}{\textbf{gpt-3.5}} & \multicolumn{3}{|c|}{\textbf{robot-gpt}}\\
\cline{2-7} 

&\textbf{Success} & \textbf{Fail} & \textbf{AP} &\textbf{Success} & \textbf{Fail} & \textbf{AP}\\ \hline
move cube(E)& 22 & 3  & 0.88& 25 & 0  & 1.0\\ \hline
block stacking(M)& 10 & 15  & 0.40& 24 & 1  & 0.96\\ \hline
pyramid stacking(M) & 8 & 17  & 0.32& 22 & 3  & 0.88\\ \hline
house building 1(M) & 12 & 13  & 0.48 & 22 & 3  & 0.88\\ \hline
house building 2(M) & 9 & 16  & 0.36 &  23 & 2  & 0.92\\ \hline
house building 3(H) & 2 & 23  & 0.08 & 20 & 5 & 0.8\\ \hline
bottle arrangement(H) & 9 & 16  & 0.36 & 23 & 2  & 0.92\\ \hline
bin packing(H) & 5 & 20  & 0.20 & 24 & 1  & 0.96\\ \hline
\end{tabular}
\label{table_result}
\end{table}

\subsection{Real robot experiment}

\begin{figure}[h]
\centering
\includegraphics[width=\linewidth]{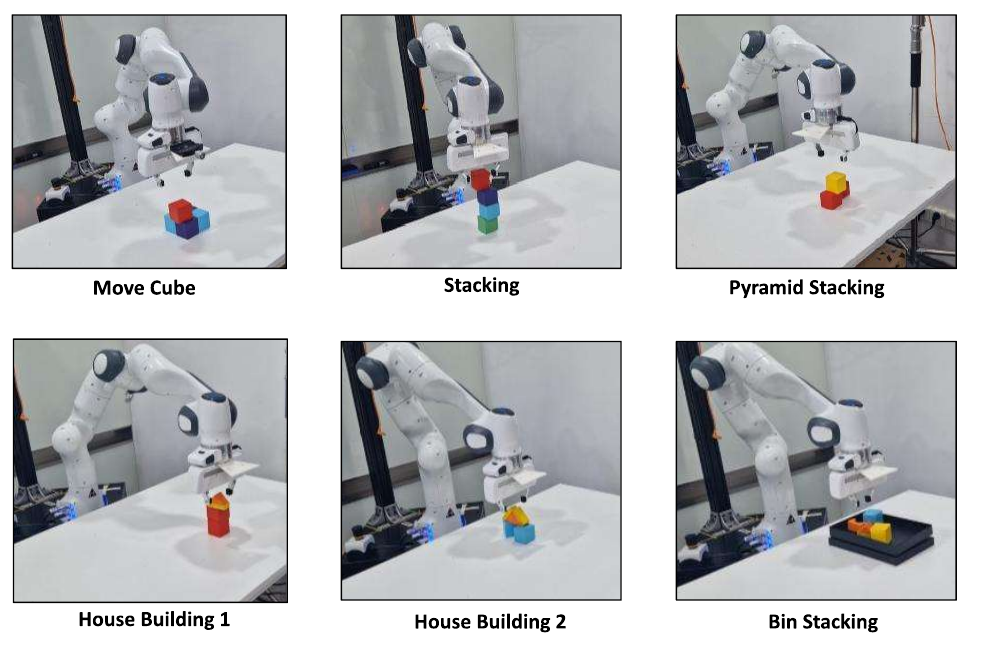}
\caption{Real robot experiments result}
\label{fig_real_res}
\end{figure}

The ultimate goal of RobotGPT is to leverage ChatGPT's intelligence to assist in solving real-world problems. Therefore, we deployed the trained agent in the real environment, which is the same as the simulation. To overcome the sim2real gap, pre-processing will be performed on the raw depth map via object segmentation and denoising before converting to the height map. Besides, to ensure a continuous process of robot pick-and-place, unlike in the simulator, the robot does not return to the observation position to capture a new depth map after its pick-up action during real robot testing. Instead, the current height map is cropped from the previous one based on the gripper holding state. For the real robot experiments, we select six scenarios that are shown in Fig \ref{fig_real_res}, and each experiment is conducted ten times. Finally, TABLE \ref{table_real_result} reports our real robot test results.

\begin{table}[h]
\centering
\caption{The counting results of real robot experiments}
\begin{tabular}{|c|c|c|c|c|}
\hline
\textbf{Task Name} & \textbf{Success} & \textbf{Fail} & \textbf{Total} &\textbf{AP} \\ \hline
move cube(E)& 8 & 2  & 10 & 0.8 \\ 
block stacking(M)& 6 & 4  & 10 & 0.6 \\
pyramid stacking(M)& 7 & 3  & 10 & 0.7 \\
house building 1(M)& 9 & 1  & 10& 0.9 \\
house building 2(M)& 6 & 4  & 10& 0.6 \\
bin packing(H)& 7 & 3  & 10 & 0.7 \\
\hline
\end{tabular}
\label{table_real_result}
\end{table}

\begin{table*}[htp]
\centering
\caption{Result of AB Test}
\begin{tabular}{|c|c|c|c|c|c|c|c|c|}
\hline
 & \multicolumn{4}{|c|}{Human Participates} & \multicolumn{4}{|c|}{RobotGPT}\\ \hline
 & CS & CQ & TU & EH & CS & CQ & TU & EH \\ \hline
move cube & 10/10 & 0.55 & 328.3 & 0/10  & \checkmark & 0.5 & 11.8 & -\\ \hline
block stacking & 10/10 & 0.72 & 199.1 & 0/10 & \checkmark & 0.75 & 44.9 & - \\ \hline
pyramid stacking & 10/10 & 0.68 & 206.6 & 0/10  & \checkmark & 0.75 & 62.7 & - \\ \hline
house building 1 & 9/10 & 0.68 & 193 & 0/9& \checkmark & 0.83 & 16.9 & - \\ \hline
house building 2 & 9/10 & 0.69 & 196 & 0/9  & \checkmark & 0.67 & 30.7 & - \\ \hline
house building 3 & 9/10 & 0.74 & 206.6 & 0/9  & \checkmark & 0.75 & 533.9 & -  \\ \hline
bottle arrangement & 7/10 & 0.74 & 575.9 & 0/7  & \checkmark & 0.86 & 110.4 & - \\ \hline
bin packing & 10/10 & 0.76 & 772.5 & 0/10  & \checkmark & 0.90& 574.8 & -  \\ \hline 
\hline
tidy up & 5/10 & 0.4374 & 1150.4 & 4/5 & \checkmark & 0.71 & 320.1 & -\\ \hline
word spell & 5/10 & 0.85 & 652 &  5/5  & \checkmark & 0.90 & 412.2 & - \\ \hline
\end{tabular}
\label{table_ab_result}
\end{table*}

From TABLE \ref{table_real_result}, it can be observed that tasks with fewer execution steps tend to have a higher number of successful attempts. The main reason for failure cases is not due to the agent's wrong predictions, but rather to insufficient precision during placement, leading to task failures. Therefore, exploring LLM in a closed-loop cycle to achieve more precise task execution would be worthwhile research in the future.

\subsection{AB Test}
\begin{figure}[h]
\centering
\includegraphics[width=\linewidth]{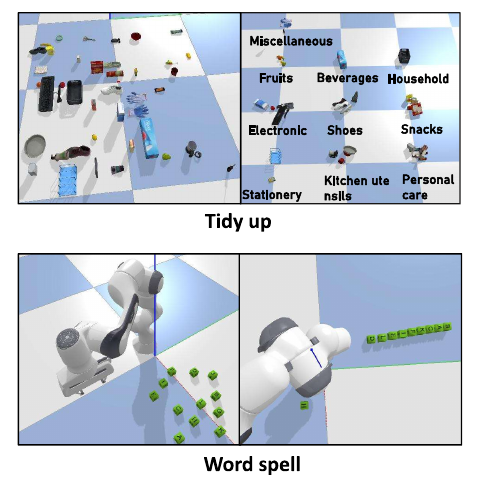}
\caption{Two open-ended experiments}
\label{fig_open_end}
\end{figure}

To investigate our LLM-driven robot's capability in solving problems that are not well-addressed by non-LLM approaches, we have introduced two open-ended experiments as Fig. \ref{fig_open_end} shown. The first experiment involves a tidy up room challenge that requires organizing 40 custom household objects, while the second one is a spelling word game that aims to spell the longest word using given set of alphabets A-L. Additionally, we invite human subjects to complete the same tasks.

\textbf{Experimental Protocol.} 
We invite ten participants for the AB test experiment. Seven of them have prior experience in robot development, while three have experience in image processing. We set a time limit of 70 minutes. Participants understand development requirements through prompts identical to those given to RobotGPT.
They are asked to finish the 10 tasks listed in Table \ref{table_ab_result} through programming.
Each participant has the autonomy to determine the order in which they attempt the tasks.

\textbf{Evaluation Metrics.} We evaluate the performance by five metrics: completion status(\textbf{CS}), for human participates, we count the number of completions, as for RobotGPT is whether finished; code quality(\textbf{CQ}) refers to the score of the generated code from 0 to 1, which is analyzed by Pylint, a static code analysis tool for Python; time usage(\textbf{TU}), consuming time in seconds from reading task requirement to implement the task in simulation; external help(\textbf{EH}) refers to whether participants search for information on the internet; For human tests, \textbf{CQ}, \textbf{TU} and \textbf{EH} represent the average value of data from individuals who have successfully completed the task.

\textbf{Results and Analysis.} 
Table \ref{table_ab_result} shows the result of AB test. Compared with hand-coding, RobotGPT demonstrates advantages in both code quality and time consumption, which are 0.762 and 221.8 seconds compared to 0.70 and 554.9 seconds for humans. Only five participants complete all the tasks within 70 minutes, therefore even for engineers with a strong programming background, generating robot demonstration data through hand-coding is time-consuming.

In addition, RobotGPT outperforms humans significantly on two open-ended tasks, tidying up room and word spelling game. This is primarily benefited from the prior knowledge repository of LLM. For the tidy-up experiment, RobotGPT divides objects into ten groups labeled as kitchenware, fruit, snacks, media, footwear, office supplies, electronics, personal care products, storage, and beverages, with 412 seconds time-consuming. This is a very satisfying result. Consider if there are 400 objects instead of 40, the advantages of LLM-driven robots would become more evident. In the spelling game, the result provided by the RobotGPT is a 9-letter word 'backfield,' while the best result of human response is a 7-letter word 'blacked'. What's more, participants need to search information online to complete two open-ended tasks, indicating that LLM possesses more comprehensive knowledge than humans.

\section{Conclusion}
In this paper, we first develop an effective prompting structure to enhance ChatGPT's understanding of the robot's environment and the tasks it needs to implement. Next, we introduce a framework called RobotGPT, which leverages ChatGPT's problem-solving capabilities to achieve more stable task execution. In experiments, we build a metric to measure task difficulty and observe that as the task difficulty increases, the success rate of execution by ChatGPT decreases.
In contrast, RobotGPT can execute these tasks with a success rate of 91.5\%, demonstrating a more stable performance. More importantly, this agent has also been deployed to run in real-world environments. Therefore, 
training a RobotGPT by utilizing ChatGPT as an expert is a more stable approach compared to directly using ChatGPT as a task planner. In addition, the AB test shows our LLM-driven robot outperforms hand-coding significantly on two open-ended tasks owing to the massive priori knowledge repository of LLM. Overall, the integration of robotics and LLMs is still at an infant stage. Our work is just an initial exploration, and we believe that much of the future research in this area is to
explore how to properly use ChatGPT’s abilities in the field of robotics. 
\bibliographystyle{ieeetr} 
\bibliography{main} %

\begin{thebibliography}{10}

\bibitem{vemprala2023chatgpt}
S.~Vemprala, R.~Bonatti, A.~Bucker, and A.~Kapoor, ``Chatgpt for robotics: Design principles and model abilities,'' {\em Microsoft Auton. Syst. Robot. Res}, vol.~2, p.~20, 2023.

\bibitem{lin2023text2motion}
K.~Lin, C.~Agia, T.~Migimatsu, M.~Pavone, and J.~Bohg, ``Text2motion: From natural language instructions to feasible plans,'' in {\em ICRA2023 Workshop on Pretraining for Robotics (PT4R)}, 2023.

\bibitem{marge2022spoken}
M.~Marge, C.~Espy-Wilson, N.~G. Ward, A.~Alwan, Y.~Artzi, M.~Bansal, G.~Blankenship, J.~Chai, H.~Daum{\'e}~III, D.~Dey, {\em et~al.}, ``Spoken language interaction with robots: Recommendations for future research,'' {\em Computer Speech \& Language}, vol.~71, p.~101255, 2022.

\bibitem{liang2022code}
J.~Liang, W.~Huang, F.~Xia, P.~Xu, K.~Hausman, B.~Ichter, P.~Florence, and A.~Zeng, ``Code as policies: Language model programs for embodied control,'' {\em arXiv preprint arXiv:2209.07753}, 2022.

\bibitem{tellex2020robots}
S.~Tellex, N.~Gopalan, H.~Kress-Gazit, and C.~Matuszek, ``Robots that use language,'' {\em Annual Review of Control, Robotics, and Autonomous Systems}, vol.~3, pp.~25--55, 2020.

\bibitem{kollar2010toward}
T.~Kollar, S.~Tellex, D.~Roy, and N.~Roy, ``Toward understanding natural language directions,'' in {\em 2010 5th ACM/IEEE International Conference on Human-Robot Interaction (HRI)}, pp.~259--266, IEEE, 2010.

\bibitem{nair2022learning}
S.~Nair, E.~Mitchell, K.~Chen, S.~Savarese, C.~Finn, {\em et~al.}, ``Learning language-conditioned robot behavior from offline data and crowd-sourced annotation,'' in {\em Conference on Robot Learning}, pp.~1303--1315, PMLR, 2022.

\bibitem{madaan2022language}
A.~Madaan, S.~Zhou, U.~Alon, Y.~Yang, and G.~Neubig, ``Language models of code are few-shot commonsense learners,'' {\em arXiv preprint arXiv:2210.07128}, 2022.

\bibitem{huang2022language}
W.~Huang, P.~Abbeel, D.~Pathak, and I.~Mordatch, ``Language models as zero-shot planners: Extracting actionable knowledge for embodied agents,'' in {\em International Conference on Machine Learning}, pp.~9118--9147, PMLR, 2022.

\bibitem{brohan2023can}
A.~Brohan, Y.~Chebotar, C.~Finn, K.~Hausman, A.~Herzog, D.~Ho, J.~Ibarz, A.~Irpan, E.~Jang, R.~Julian, {\em et~al.}, ``Do as i can, not as i say: Grounding language in robotic affordances,'' in {\em Conference on Robot Learning}, pp.~287--318, PMLR, 2023.

\bibitem{wu2023tidybot}
J.~Wu, R.~Antonova, A.~Kan, M.~Lepert, A.~Zeng, S.~Song, J.~Bohg, S.~Rusinkiewicz, and T.~Funkhouser, ``Tidybot: Personalized robot assistance with large language models,'' {\em arXiv preprint arXiv:2305.05658}, 2023.

\bibitem{kroemer2021review}
O.~Kroemer, S.~Niekum, and G.~Konidaris, ``A review of robot learning for manipulation: Challenges, representations, and algorithms,'' {\em The Journal of Machine Learning Research}, vol.~22, no.~1, pp.~1395--1476, 2021.

\bibitem{cabi2019scaling}
S.~Cabi, S.~G. Colmenarejo, A.~Novikov, K.~Konyushkova, S.~Reed, R.~Jeong, K.~Zolna, Y.~Aytar, D.~Budden, M.~Vecerik, {\em et~al.}, ``Scaling data-driven robotics with reward sketching and batch reinforcement learning,'' {\em arXiv preprint arXiv:1909.12200}, 2019.

\bibitem{wang2022equivariant}
D.~Wang, R.~Walters, X.~Zhu, and R.~Platt, ``Equivariant $ q $ learning in spatial action spaces,'' in {\em Conference on Robot Learning}, pp.~1713--1723, PMLR, 2022.

\bibitem{zeng2022robotic}
A.~Zeng, S.~Song, K.-T. Yu, E.~Donlon, F.~R. Hogan, M.~Bauza, D.~Ma, O.~Taylor, M.~Liu, E.~Romo, {\em et~al.}, ``Robotic pick-and-place of novel objects in clutter with multi-affordance grasping and cross-domain image matching,'' {\em The International Journal of Robotics Research}, vol.~41, no.~7, pp.~690--705, 2022.

\bibitem{lee2021ikea}
Y.~Lee, E.~S. Hu, and J.~J. Lim, ``Ikea furniture assembly environment for long-horizon complex manipulation tasks,'' in {\em 2021 ieee international conference on robotics and automation (icra)}, pp.~6343--6349, IEEE, 2021.

\bibitem{zhu2020robosuite}
Y.~Zhu, J.~Wong, A.~Mandlekar, R.~Mart{\'\i}n-Mart{\'\i}n, A.~Joshi, S.~Nasiriany, and Y.~Zhu, ``robosuite: A modular simulation framework and benchmark for robot learning,'' {\em arXiv preprint arXiv:2009.12293}, 2020.

\bibitem{delhaisse2020pyrobolearn}
B.~Delhaisse, L.~Rozo, and D.~G. Caldwell, ``Pyrobolearn: A python framework for robot learning practitioners,'' in {\em Conference on Robot Learning}, pp.~1348--1358, PMLR, 2020.

\bibitem{james2020rlbench}
S.~James, Z.~Ma, D.~R. Arrojo, and A.~J. Davison, ``Rlbench: The robot learning benchmark \& learning environment,'' {\em IEEE Robotics and Automation Letters}, vol.~5, no.~2, pp.~3019--3026, 2020.

\bibitem{wang2023bulletarm}
D.~Wang, C.~Kohler, X.~Zhu, M.~Jia, and R.~Platt, ``Bulletarm: An open-source robotic manipulation benchmark and learning framework,'' in {\em Robotics Research}, pp.~335--350, Springer, 2023.

\bibitem{openai}
OpenAI, ``Best practices for prompt engineering with openai api,'' 8 2023.
\newblock Accessed: August 23, 2023.

\bibitem{wang2020policy}
D.~Wang, C.~Kohler, and R.~Platt, ``Policy learning in se (3) action spaces,'' {\em arXiv preprint arXiv:2010.02798}, 2020.

\bibitem{mnih2015human}
V.~Mnih, K.~Kavukcuoglu, D.~Silver, A.~A. Rusu, J.~Veness, M.~G. Bellemare, A.~Graves, M.~Riedmiller, A.~K. Fidjeland, G.~Ostrovski, {\em et~al.}, ``Human-level control through deep reinforcement learning,'' {\em nature}, vol.~518, no.~7540, pp.~529--533, 2015.

\bibitem{lakshminarayanan2016reinforcement}
A.~S. Lakshminarayanan, S.~Ozair, and Y.~Bengio, ``Reinforcement learning with few expert demonstrations,'' in {\em NIPS workshop on deep learning for action and interaction}, vol.~2016, 2016.

\bibitem{hester2018deep}
T.~Hester, M.~Vecerik, O.~Pietquin, M.~Lanctot, T.~Schaul, B.~Piot, D.~Horgan, J.~Quan, A.~Sendonaris, I.~Osband, {\em et~al.}, ``Deep q-learning from demonstrations,'' in {\em Proceedings of the AAAI Conference on Artificial Intelligence}, vol.~32, 2018.

\bibitem{zeng2018learning}
A.~Zeng, S.~Song, S.~Welker, J.~Lee, A.~Rodriguez, and T.~Funkhouser, ``Learning synergies between pushing and grasping with self-supervised deep reinforcement learning,'' in {\em 2018 IEEE/RSJ International Conference on Intelligent Robots and Systems (IROS)}, pp.~4238--4245, IEEE, 2018.

\bibitem{long2015fully}
J.~Long, E.~Shelhamer, and T.~Darrell, ``Fully convolutional networks for semantic segmentation,'' in {\em Proceedings of the IEEE conference on computer vision and pattern recognition}, pp.~3431--3440, 2015.

\end{thebibliography}



\end{document}